\documentclass[sigconf,9pt]{acmart}

\usepackage{booktabs}
\usepackage{array}
\usepackage{graphicx}
\usepackage{subcaption}
\usepackage{multirow}
\usepackage{siunitx}
\usepackage{xcolor}
\usepackage{needspace}

\title{Agentic Performance at the Edge: Insights from Benchmarking}

\author{Shiqiang Wang}
\affiliation{%
  \institution{University of Exeter}
  \city{Exeter}
  \country{United Kingdom}}
\email{s.wang9@exeter.ac.uk}
\author{Herbert Woisetschl\"ager}
\affiliation{%
  \institution{Technical University of Munich}
  \city{Munich}
  \country{Germany}}
\email{h.woisetschlaeger@tum.de}

\copyrightyear{2026}
\acmYear{2026}
\setcopyright{cc}
\setcctype{by}
\acmConference[MobiSys Workshop '26]{The 24th Annual International Conference on Mobile Systems, Applications and Services}{June 21--25, 2026}{Cambridge, United Kingdom}
\acmBooktitle{The 24th Annual International Conference on Mobile Systems, Applications and Services (MobiSys Workshop '26), June 21--25, 2026, Cambridge, United Kingdom}
\acmDOI{10.1145/3812836.3814741}
\acmISBN{979-8-4007-2712-2/2026/06}


\usepackage{float}


\begin{document}

\begin{abstract}
Agentic artificial intelligence (AI) is a natural fit for Internet of Things (IoT) and edge systems, but edge deployments are often constrained to models around 8 billion parameters or smaller.
An important question is: How much agentic-task quality is lost when model size is constrained by memory, power, and latency budgets? 
To address this question, in this paper, we provide an initial empirical study considering edge-focused model scaling, general-purpose versus coder-oriented model effects, and tool-enabled execution under a fixed protocol. 
We introduce a domain-conditioned evaluation methodology, an implementation-grounded analysis of model-tool interactions, practical guidance for model selection under constraints, and an analysis of failure modes that reveals distinct semantic versus execution failure patterns across model families. 
Our core finding is that edge-agent quality is not a simple function of parameter count. Robust deployment depends on the joint design of model choice and tool workflow. Domain-conditioned analysis reveals Pareto fronts in the accuracy-latency space that can guide strategy selection based on operational priorities.
\end{abstract}

\maketitle

\section{Introduction}
Agentic large language model (LLM) systems have progressed from single-turn generation to iterative workflows that reason, plan, and act with tools~\cite{yao2023react,madaan2023selfrefine}. 
This shift is strategically important for edge and near-edge environments because local agents can reduce cloud round-trips, preserve private data, and continue operating during intermittent connectivity. 
In practice, however, edge reliability is bounded by strict compute and latency budgets. Local models must be small enough to run consistently, but accurate enough to support operational decisions. 
Particularly in industrial and enterprise edge settings, reliability, auditability, and service continuity are primary system goals.

This creates a core systems tension. 
If model quality scales smoothly with parameter count, model selection is straightforward. In that case, we can simply run the largest model that fits. 
If not, parameter count is an incomplete control variable, and architecture-level decisions, such as verification, fallback, and selective offload, become important. 
The challenge is amplified by the hardware heterogeneity of practical edge systems. 
Production edge deployments range from on-premise gateways and compact edge servers to private micro-datacenter nodes and CPU-only industrial controllers. 
On many of such platforms, models above a certain size can trigger unacceptable latency tails or memory pressure, even if the average throughput is acceptable. Moreover, in operational diagnostics, an agent must fuse multiple evidence sources and produce concise machine-actionable outputs. 
Wrong predictions can trigger noisy escalations or delayed remediation. 
For all these reasons, model selection of edge agents is much more complex than a simple ``best benchmark score'' under hard resource constraints~\cite{nvidia2024jetsonnano,yuan2025localcloud}.

Recent compact-model progress (e.g., TinyLlama and Phi-3) indicates that smaller models can remain useful under constrained hardware~\cite{zhang2024tinyllama,abdin2024phi3}. 
At the same time, prior agentic benchmark studies (e.g., AgentBench, GAIA) show interactive tool-grounded tasks remain significantly harder than static question answering (QA)~\cite{liu2024agentbench,mialon2023gaia}.
In addition, practical edge deployments are often limited in usable context window (commonly much smaller than frontier-scale settings), which further motivates evaluating model behavior under constrained runtime settings.
We build on these insights with a workload-aligned evaluation that emphasizes operational diagnostics and deployment decisions instead of leaderboard ranking.

In this paper, we evaluate controlled comparisons under one consistent runtime stack, including size scaling, generation comparison at similar capacity, and general-purpose versus coder-oriented model variants. 
We intentionally avoid per-model prompt tuning, so that we can preserve comparability, keep broad practical applicability, and expose transferability challenges between models and agent runtimes. 
Beyond accuracy, we evaluate failure composition as a primary metric, i.e., understanding whether failures stem from semantic errors (wrong answers despite completing the trajectory) or execution errors (tool failures or step limits), which is important for operational strategy selection.

Our main contributions are as follows: 
\begin{enumerate}
    \item A full edge-style agent evaluation with consistent tool workflow and scoring.
    \item A controlled decomposition of size, generation, and variant effects across multiple open model ecosystems.
    \item An analysis showing that domain asymmetry can dominate model-size differences in deployment settings.
    \item A failure-mode analysis conditioned on task domain (FinOps vs SRE, see Section~\ref{sec:benchmark_setup} for their definitions), revealing distinct semantic versus execution failure patterns across model families, inspiring targeted improvement strategies.
\end{enumerate}

\section{Related Work}
\textbf{Agentic benchmarks.} There exist multiple benchmarks for evaluating agentic AI systems. For example, GAIA evaluates broad assistant competence on heterogeneous multi-step tasks~\cite{mialon2023gaia}, while AgentBench focuses on interactive environments and agent execution traces~\cite{liu2024agentbench}. OfficeBench targets realistic office automation workflows~\cite{wang2024officebench}, and workload-aligned benchmarks like ITBench shift toward incident diagnosis and automation tasks requiring root-cause localization under operational constraints~\cite{jha2025itbench}. 

\textbf{Reasoning-loop design.} ReAct and self-refinement motivate iterative action-reflection loops~\cite{yao2023react,madaan2023selfrefine}. Chain-of-thought and zero-shot reasoning show that outcome quality is sensitive to inference structure~\cite{wei2022cot,kojima2022zeroshot}. In our work, we hold the tool workflow fixed across models, so that differences are attributable primarily to model capability instead of prompt drift.

\textbf{Efficiency and deployment constraints.} Compact model families and quantization improve feasibility on constrained hardware~\cite{zhang2024tinyllama,abdin2024phi3,dettmers2023qlora}. However, deployment feasibility does not imply monotonic quality scaling in tool-enabled settings. Hybrid local-cloud inference has been shown to improve quality-latency-cost trade-offs~\cite{yuan2025localcloud}, and frontier model reports show broad capability gains at scale~\cite{openai2023gpt4}, but transfer to constrained tool-heavy tasks is not guaranteed.

Different from these related works, we systematically evaluate the performance of agentic tasks at the edge in this paper, which has not been studied in existing literature.

\section{Methodology}
\subsection{Research Questions}
\label{sec:research_questions}
We evaluate three deployment-facing questions as follows.
\begin{itemize}
    \item \textbf{RQ1 (Size):} How does agentic-task reliability vary across model sizes under edge constraints?
    \item \textbf{RQ2 (Generation):} At similar capacity, do newer model generations reliably improve outcomes?
    \item \textbf{RQ3 (Variant):} At matched sizes, when do coder-oriented models improve over general-purpose models?
\end{itemize}

\subsection{Benchmark Setup and Design Principles}
\label{sec:benchmark_setup}
Our methodology is designed to transfer across domains and is not tied to a specific benchmark. The key design idea is \emph{controlled decomposition}, studying model size effects, generation effects, and general-purpose versus coder-oriented effects with one fixed tool workflow, then interpreting outcomes by domain and failure type.

Specifically, we instantiate this design on an operational benchmark with two diagnostic domains, including financial operations (FinOps) and site reliability engineering (SRE), both of which are from ITBench~\cite{jha2025itbench}. We use the scorable task subset in our environment and hold the task ordering and seed protocol fixed. It is worth noting that the benchmark choice here matters less than the evaluation discipline, where the latter involves the same tasks, same tool protocol, same scorer, and same runtime settings across all model families.

This framing avoids a common confound in agent papers involving changing prompts, tool semantics, and stopping rules at the same time as changing models. By separating these factors, model comparisons become more interpretable and more useful for deployment decisions.

\subsection{Tool and Action-Space Design}

The important design choice is not any single tool, but the \emph{shape of the decision loop}. A robust loop should let the model 1) gather evidence, 2) test alternatives, and 3) finalize with a compact machine-checkable answer. Most surprising errors in our runs emerge when one of these stages collapses. For example, plausible evidence is retrieved, but candidate reduction or finalization fails.

We therefore treat tool design as a systems interface problem. 
A key consideration is that, if the action space is too narrow, the model cannot resolve ambiguity. If it is too broad, trajectory variance and error cascades increase. 
Thus, in our design, the tool layer is a minimal but expressive action space. It includes retrieval, topology/context expansion, candidate reduction, and domain-specific analyzers, as shown in Table~\ref{tab:itops_tools}. This structure is general and maps to many enterprise agent settings beyond IT operations.

\begin{table}[t]
\caption{ITOps tool families used in this evaluation stack.}
\label{tab:itops_tools}
\centering
\small
\begin{tabular}{lp{0.60\linewidth}}
\toprule
\textbf{Category} & \textbf{Representative tools in harness} \\
\midrule
Observability retrieval & Log queries, trace queries, alert retrieval \\
Topology reasoning & Node lookup, neighborhood traversal, path checks \\
Candidate reduction & Candidate summarization and ranking \\
Cost anomaly analysis & Structured anomaly decomposition \\
\bottomrule
\end{tabular}
\vspace{-1em}
\end{table}

The system with the tools given in Table~\ref{tab:itops_tools} is a staged evidence-fusion setup instead of a single-tool question answering setting. This component view enables more useful interpretation than aggregate accuracy alone, because the same final error can originate from retrieval, ranking, or finalization.

\subsection{Inference Protocol}
We run one fixed tool-enabled iterative protocol for all models. 
Final answers are scored by the ITBench root-cause evaluator under identical post-processing and runtime settings.
We use a 12-step limit as a conservative compute budget, which is consistent with common benchmark practice and suitable for constrained edge hardware~\cite{mialon2023gaia,liu2024agentbench}.
To reduce confounding, the harness enforces one tool call per turn with a matching tool result before the next model action. This avoids mixed multi-tool responses and keeps trace semantics comparable across models.

\subsection{Execution Environment}
All models are served via the same local inference endpoint on an NVIDIA RTX Pro 6000 Blackwell GPU. Our intended deployment context is industrial and enterprise edge operations. The local serving setup should be interpreted as a controlled proxy testbed for reproducible comparison, instead of the target production hardware. We keep tool workflow, timeout settings, and scorer fixed across experiments to isolate model-family effects from infrastructure confounders.

\subsection{Reproducibility Controls}
Our benchmark runner records the selected task list, per-task traces, and structured summary outputs for each run. This makes post-hoc validation straightforward, since every reported aggregate can be recomputed from persisted run artifacts. We also keep the benchmark following the aforementioned protocol of single tool call followed by single tool result, to avoid hidden protocol drift across models.

We use a consistent runtime profile across experiments, with 25 FinOps tasks and 35 SRE tasks that have root-cause ground truth labels available for accuracy computation, along with tool mode enabled, fixed step limit, fixed model-level parallelism setting, and an inference temperature of 0.2. Each model is evaluated with three different random seeds, and reported plots and tables use mean values with standard-deviation estimates. This approach ensures statistical robustness while maintaining comparability across the model ecosystem.

We also \textit{release the code for our experiments} at \url{https://github.com/core-aix/agentic-edge-insights}. The codebase includes more details about the tool call procedure and experimental setup in general, in addition to enabling the community to reproduce our experiments and possibly extend the code to include other benchmarks.

\subsection{Model Sets and Experimental Setup}
In our current experiments, we use three representative open-model ecosystems (Qwen, Phi, and Mistral) to test whether observed patterns repeat across families.\footnote{We intentionally compare a limited number of open ecosystems under a fixed stack to isolate effects of size, generation, or model-variant. As we only use a specific benchmark (ITBench) with a specific protocol (including prompt template) in this paper, the result \textit{should not} be interpreted as a universal ranking of model families.} Each experiment isolates one variable (size, generation, or model variant type) at a time. 

We evaluate our three deployment-facing questions (as described in  Section~\ref{sec:research_questions}). These questions directly address the concerns of practitioners deploying agents in resource-constrained environments, where the choice of model can significantly impact operational reliability, latency, cost-effectiveness, and overall system maintainability. By answering these questions, we provide empirical evidence to guide model selection decisions in production edge deployments and help organizations make informed choices about model deployment strategies.

\subsection{Illustrative Agent Trace}
To illustrate the concrete behavior of the agent, we include a \textit{narrated} successful trajectory from a FinOps anomaly task as follows: 
\begin{itemize}
    \item \textbf{User task:} Identify the root-cause service for a cloud-cost anomaly.
    \item \textbf{Assistant step 1:} Call the \texttt{analyze\_finops\_cost\_anomaly} tool with scenario data, anomaly date, and account identifier.
    \item \textbf{Tool result 1:} Top candidates include \texttt{S3} and \texttt{Lambda}, with evidence tied to the anomaly window.
    \item \textbf{Assistant step 2:} Call the \texttt{analyze\_finops\_cost\_anomaly} tool again with a narrowed window and candidate-focused filters for validation.
    \item \textbf{Tool result 2:} Validation confirms \texttt{S3} as the dominant contributor in the anomaly interval.
    \item \textbf{Assistant final answer:} Provide the final answer: \texttt{S3}.
\end{itemize}
This trace illustrates why protocol constraints matter. One tool call per turn and clear tool-result grounding make trajectories auditable and reduce hidden prompt effects in cross-model comparisons.

\section{Results}
\subsection{Overall Performance}
\begin{figure}[b]
    \centering
    \begin{subfigure}[t]{0.32\linewidth}
        \centering
        \includegraphics[width=\linewidth]{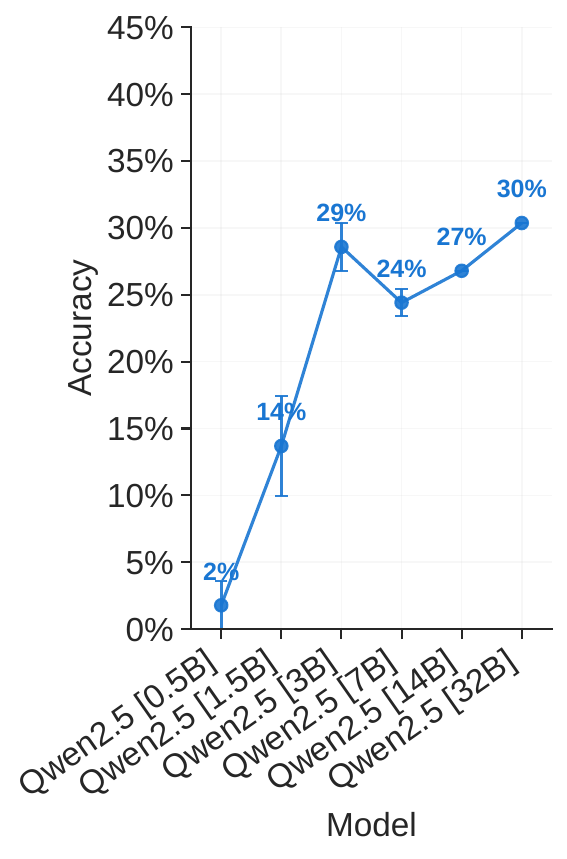}\vspace{-0.5em}
        \caption{Qwen}
    \end{subfigure}
    \begin{subfigure}[t]{0.32\linewidth}
        \centering
        \includegraphics[width=\linewidth]{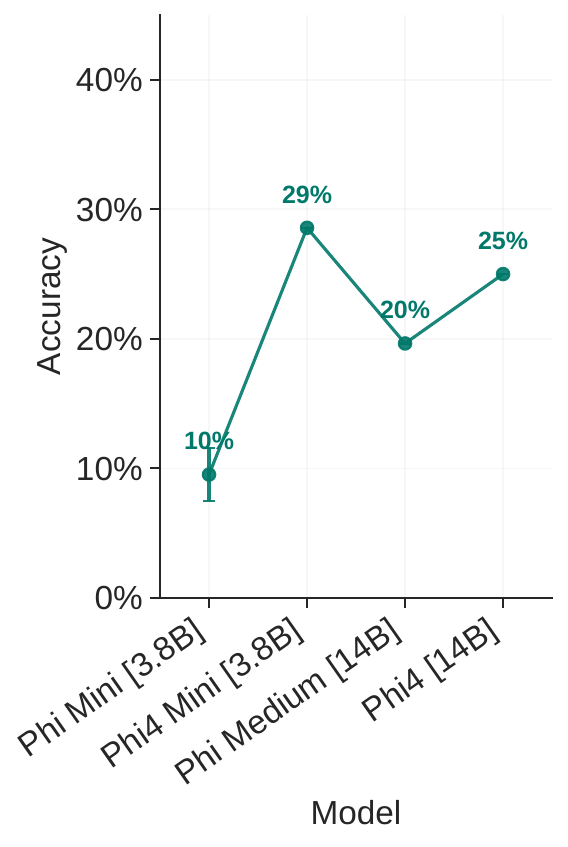}\vspace{-0.5em}
        \caption{Phi}
    \end{subfigure}
    \begin{subfigure}[t]{0.32\linewidth}
        \centering
        \includegraphics[width=\linewidth]{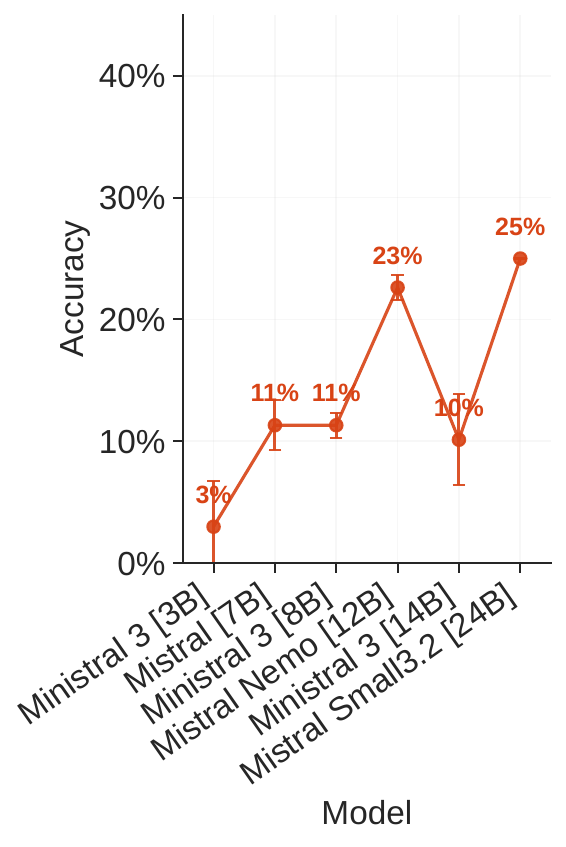}\vspace{-0.5em}
        \caption{Mistral}
    \end{subfigure}
    \caption{Size-view accuracy by family (each subfigure uses the family's own ascending size order).}
    \label{fig:acc_latest}
\end{figure}

\begin{figure}[t]
    \centering
    \begin{subfigure}[t]{0.32\linewidth}
        \centering
        \includegraphics[width=\linewidth]{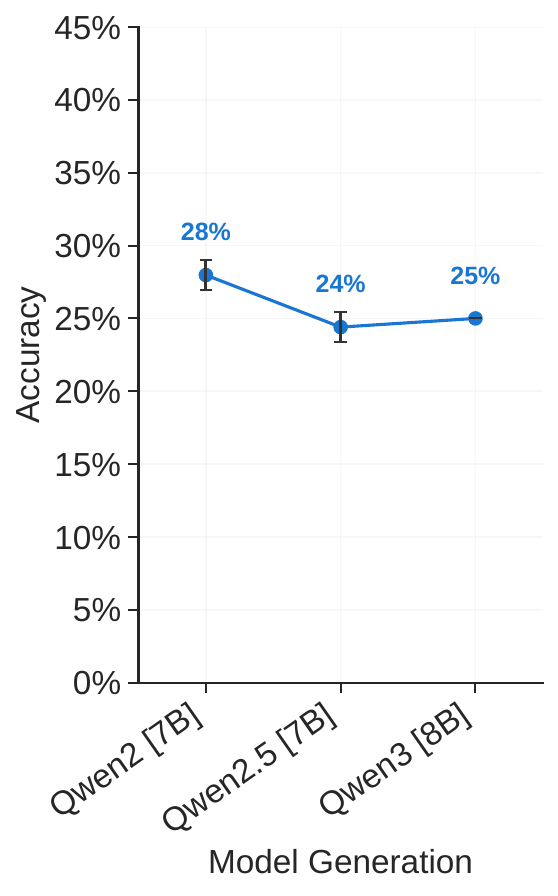}\vspace{-0.5em}
        \caption{Qwen}
    \end{subfigure}
    \begin{subfigure}[t]{0.32\linewidth}
        \centering
        \includegraphics[width=\linewidth]{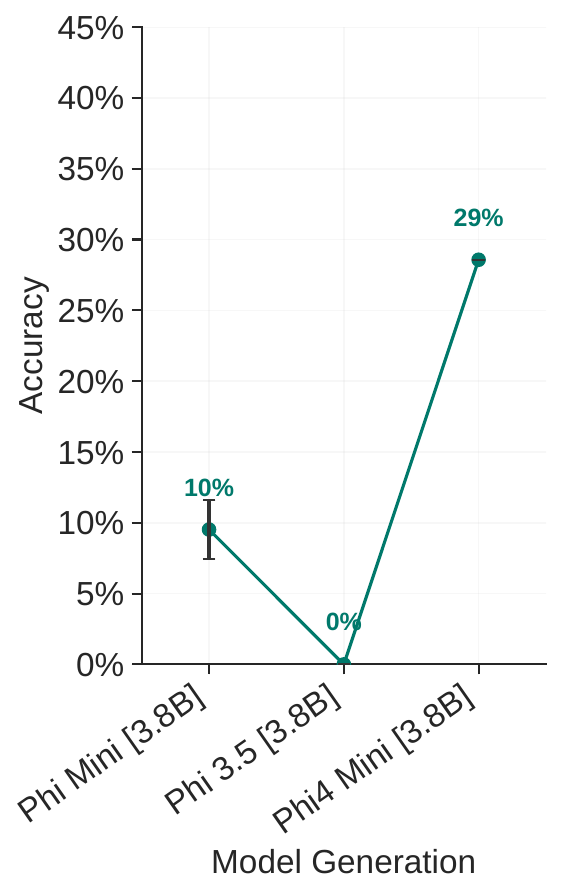}\vspace{-0.5em}
        \caption{Phi}
    \end{subfigure}
    \begin{subfigure}[t]{0.32\linewidth}
        \centering
        \includegraphics[width=\linewidth]{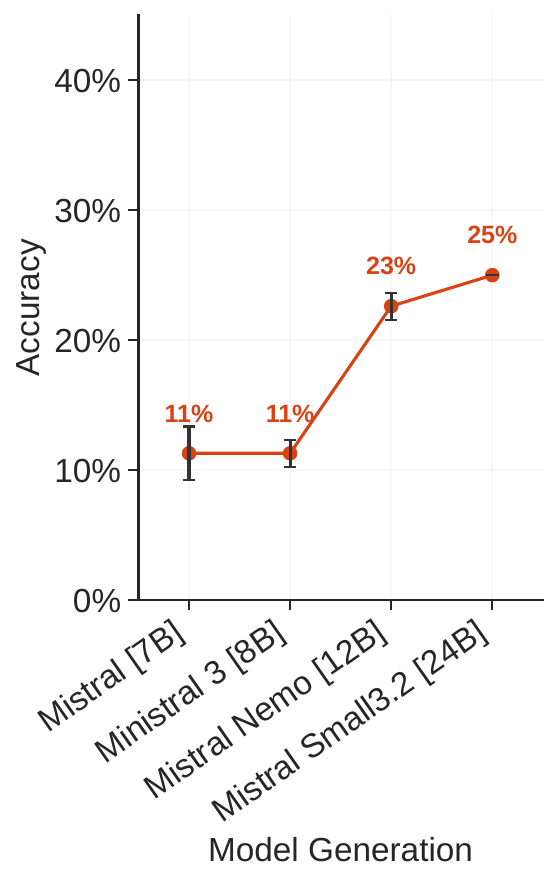}\vspace{-0.5em}
        \caption{Mistral}
    \end{subfigure}
    \caption{Generation-view accuracy by family.}
    \label{fig:acc_versions}
\end{figure}

As shown in the results in Figures~\ref{fig:acc_latest}--\ref{fig:tradeoff_scatters}, the same pattern can be observed across all the experiments. Namely, model variant type and generation changes can help, but improvements are uneven and not strictly steady with size. 

Several observations are especially relevant for deployment decisions. First, the strongest overall scores come from Qwen coder-oriented variants (33.9\% mean accuracy), but they occupy different latency points. The 7B coder variant reaches the same top accuracy as larger coder variants at much lower latency. Second, in Phi, the best configuration reaches 28.6\% while a second strong option is 25.0\%, and their latency profiles differ substantially. Third, the strongest Mistral-family point reaches 25.0\%, remaining below the strongest Qwen and Phi points and showing a wider spread across models in the same family.

It is also important to interpret these outcomes against task difficulty. A public leaderboard for this benchmark suggests that the strongest scores are typically obtained by substantially larger models that are often proprietary~\cite{kaggle2026itbench}. With this in mind, the performance in this study is meaningful as an edge-deployment signal instead of a leaderboard-competitive result.

\subsection{Size and Generation Trends}

Figure~\ref{fig:acc_latest} shows a non-steady size curve in all three families. Gains appear with size increases, but larger models do not uniformly dominate mid-size models. In practical terms, these curves reject a simple ``largest model wins'' rule and motivate more sophisticated model selection strategies.

Figure~\ref{fig:acc_versions} indicates that generation upgrades can help, but improvements remain uneven and latency-sensitive. In Qwen, the 7B-class trend is not steady. In Phi, we observe both a sharp improvement (to Phi4 Mini 3.8B) and a collapse case (Phi 3.5 3.8B near 0\%), implying that generation labels alone do not guarantee better agentic outcomes.

\subsection{Domain-Conditioned Analysis}
\begin{figure}[t]
    \centering
    \begin{subfigure}[t]{0.32\linewidth}
        \centering
        \includegraphics[width=\linewidth]{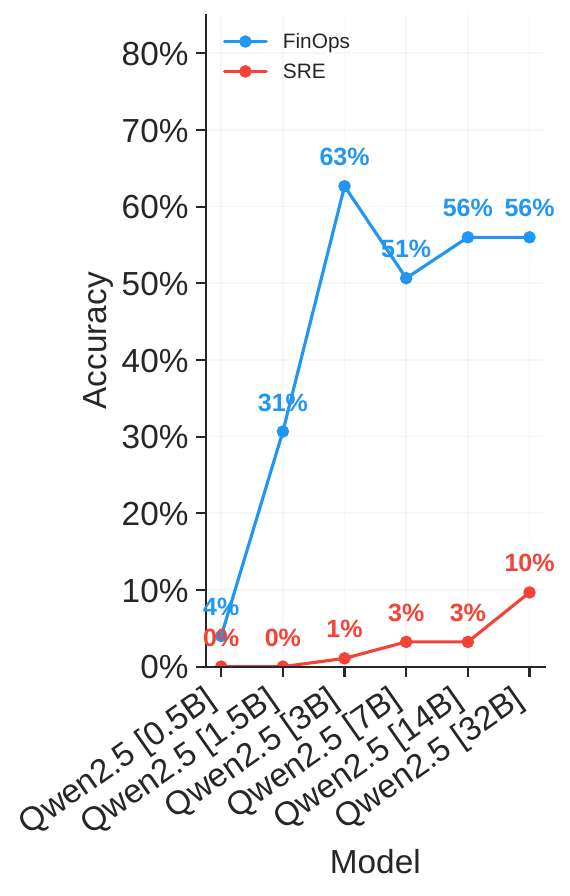}\vspace{-0.5em}
        \caption{Qwen}
    \end{subfigure}
    \begin{subfigure}[t]{0.32\linewidth}
        \centering
        \includegraphics[width=\linewidth]{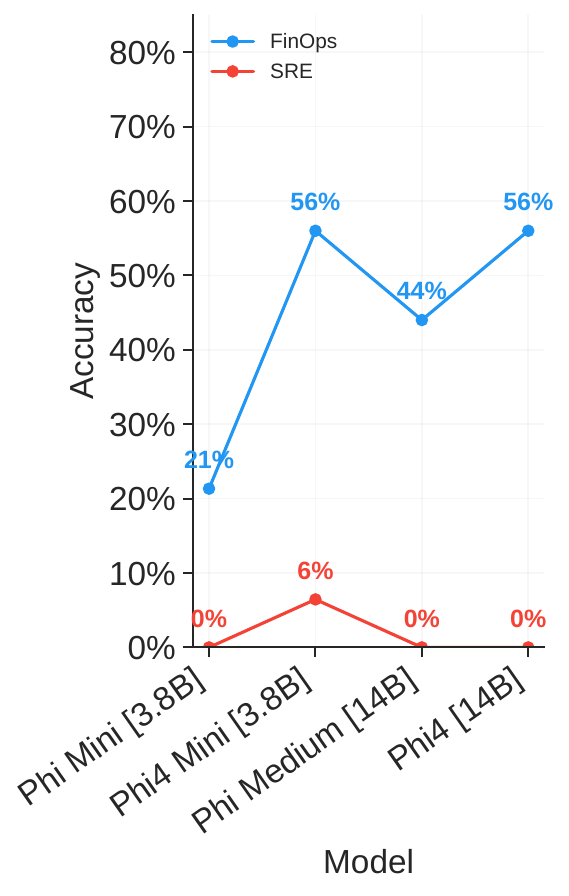}\vspace{-0.5em}
        \caption{Phi}
    \end{subfigure}
    \begin{subfigure}[t]{0.32\linewidth}
        \centering
        \includegraphics[width=\linewidth]{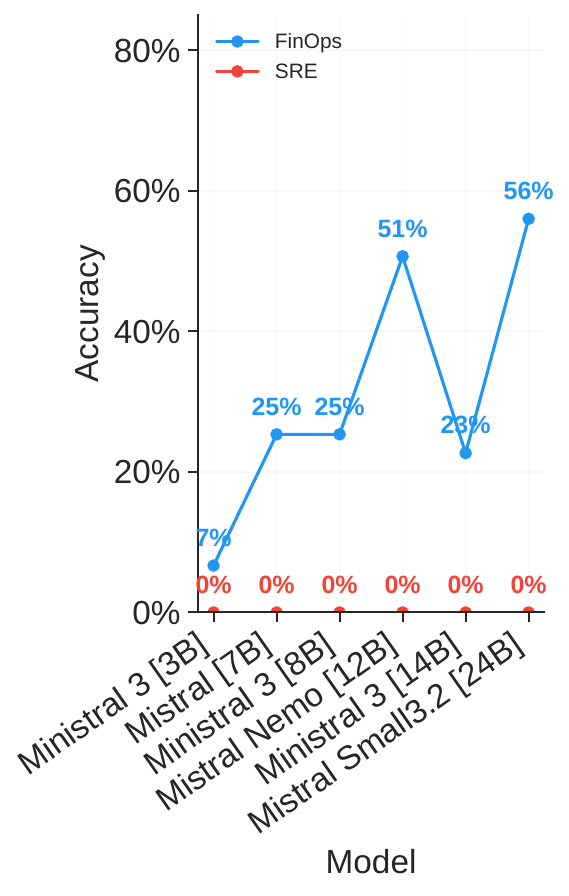}\vspace{-0.5em}
        \caption{Mistral}
    \end{subfigure}
    \caption{Domain split (FinOps vs SRE) in the size view, shown per family.}
    \label{fig:size_diff}
\end{figure}

Domain asymmetry is clearly observed in Figure~\ref{fig:size_diff}. FinOps tasks are consistently easier than SRE tasks across all the experiments. This remains true regardless of whether we vary size, generation, or model variant type, indicating that domain structure is a dominant factor in edge-agent reliability.

The FinOps domain involves cost anomaly detection and root-cause attribution, which typically requires analyzing structured financial data and identifying patterns in billing or usage records. SRE tasks, in contrast, involve multi-source log aggregation, topology reasoning, and root-cause localization in distributed systems. They are tasks that require combining heterogeneous evidence streams under uncertainty. Due to this difference, we observe that some models perform worse on SRE although they do better on FinOps, as models can be good in one type of evidence compared to the other depending on the data they were trained on.

We also note that the gap between FinOps and SRE is large. Top FinOps values are around 56--63\% depending on family and experiment, while top SRE values are much lower (roughly 6--10\%). This separation is larger than many within-family size differences, so domain mix is an important factor to consider for edge-agent deployment.

\subsection{Variant and Cross-Family Comparison}

\begin{figure}[t]
    \centering
    \begin{subfigure}[t]{0.48\linewidth}
        \centering
        \includegraphics[width=\linewidth]{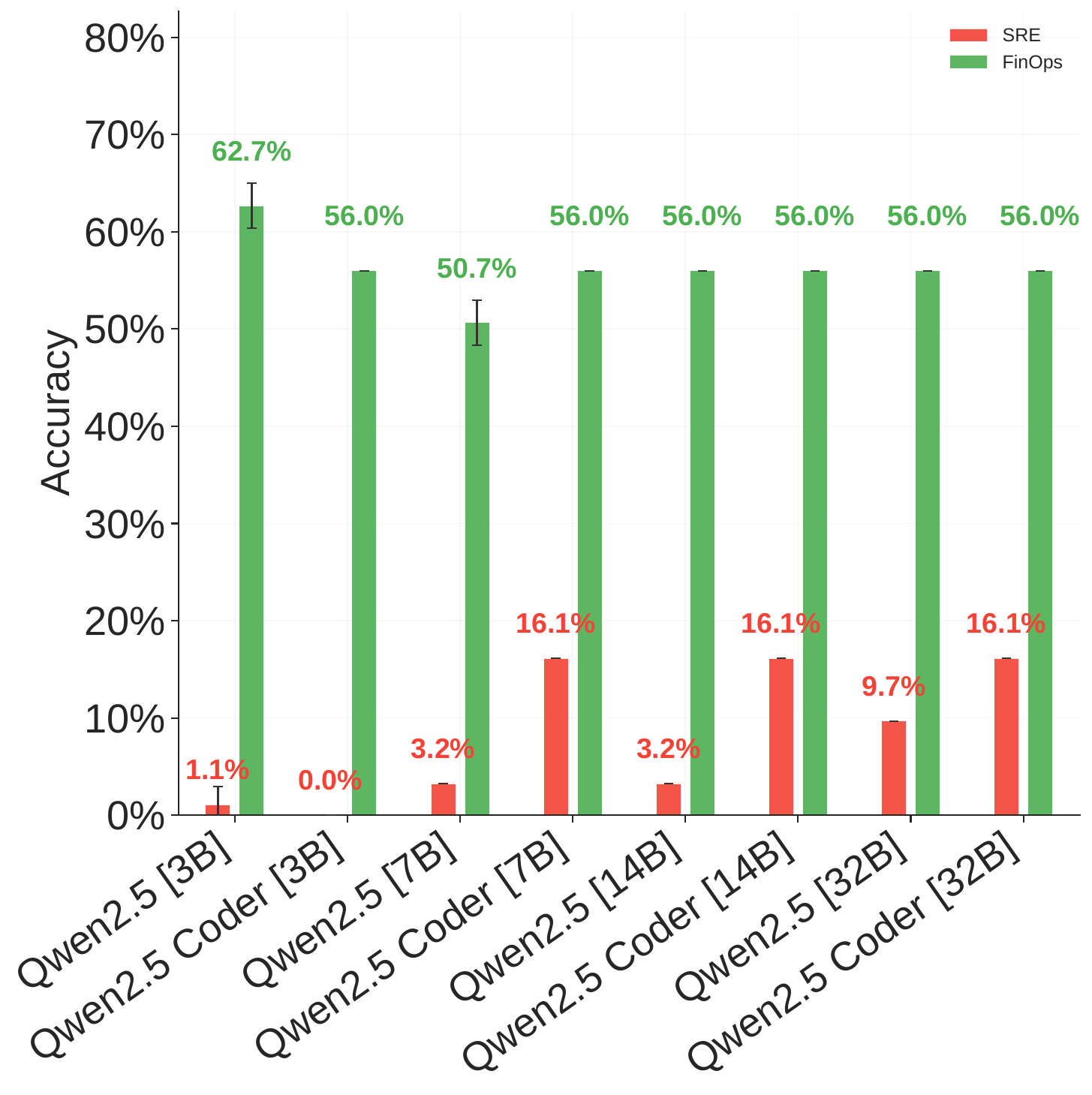}\vspace{-0.5em}
        \caption{Qwen}
    \end{subfigure}
    \begin{subfigure}[t]{0.48\linewidth}
        \centering
        \includegraphics[width=\linewidth]{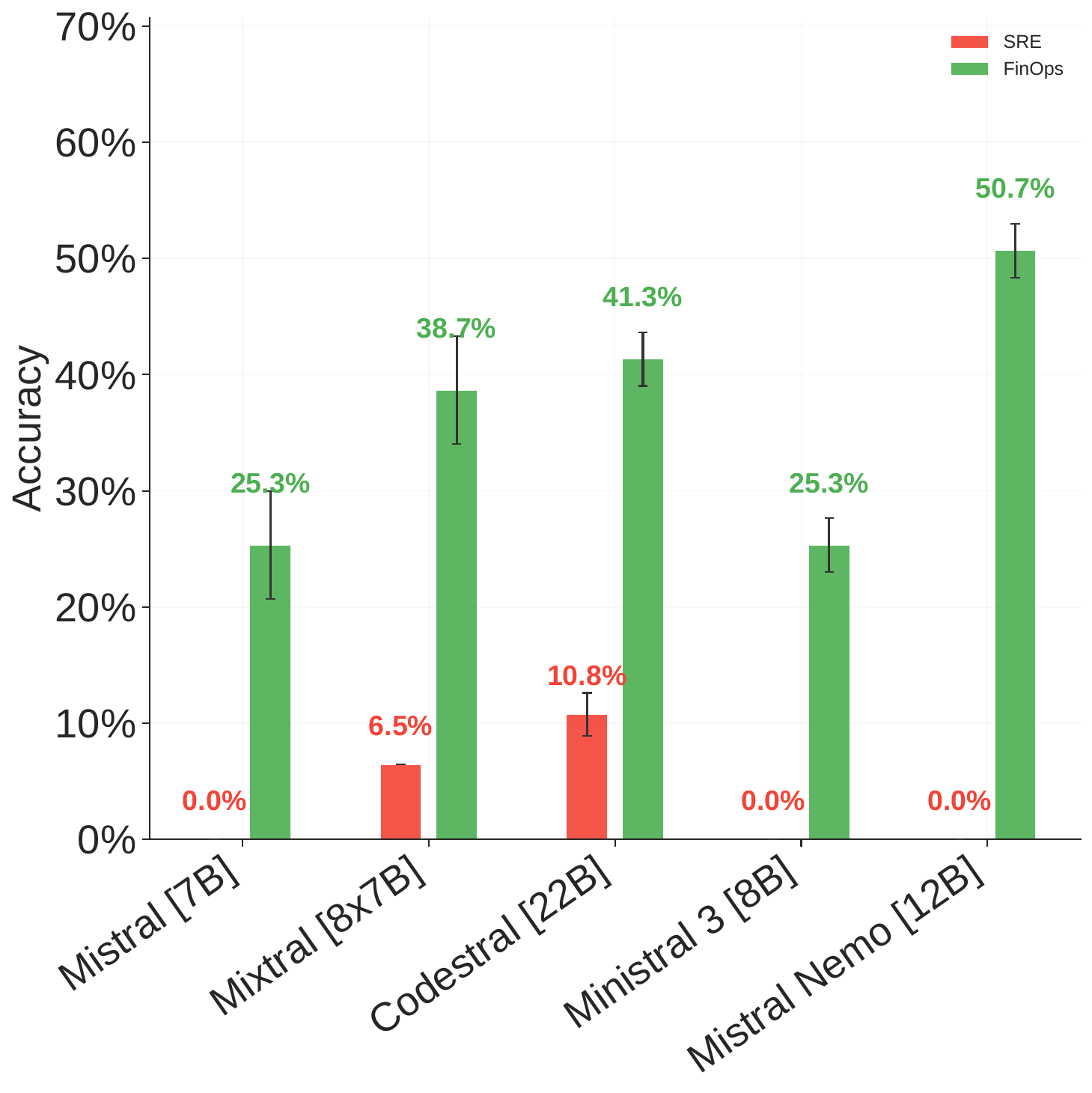}\vspace{-0.5em}
        \caption{Mistral}
    \end{subfigure}
    \caption{Per-family variant view including all relevant points in each ecosystem (Mistral uses a curated variant set: Mistral, Mixtral, Codestral, Ministral, and Mistral Nemo).}
    \label{fig:variant_all}
    \vspace{1em}
\end{figure}

From Figure~\ref{fig:variant_all}, we observe that variant effects are not uniform across scales. In Qwen, the coder-minus-general-purpose gain is strongly size-dependent, i.e., negligible or negative at 3B, large at 7B, and still positive at 14B/32B. Importantly, much of this gain comes from better SRE performance instead of only FinOps uplift, suggesting coder-oriented models help most when the task requires stronger evidence resolution.
The Mistral family shows similar non-steady behavior, with coder-oriented variants showing threshold-like improvements at specific capacity levels.

\begin{figure*}[t]
    \centering
    \begin{subfigure}[t]{0.43\textwidth}
        \centering
        \includegraphics[width=\linewidth]{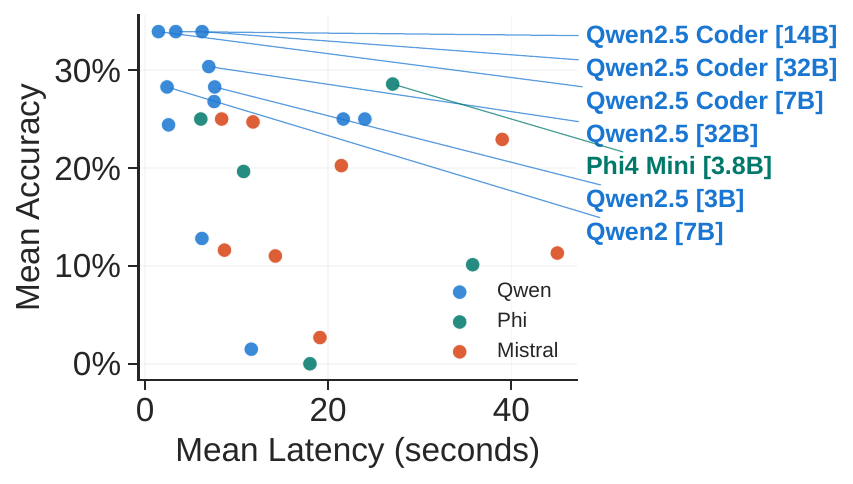}\vspace{-0.5em}
        \caption{Latency-accuracy scatter across Qwen, Phi, and Mistral models, with model-name annotations for each point.}
        \label{fig:latency_scatter}
    \end{subfigure}
    \hfill
    \begin{subfigure}[t]{0.43\textwidth}
        \centering
        \includegraphics[width=\linewidth]{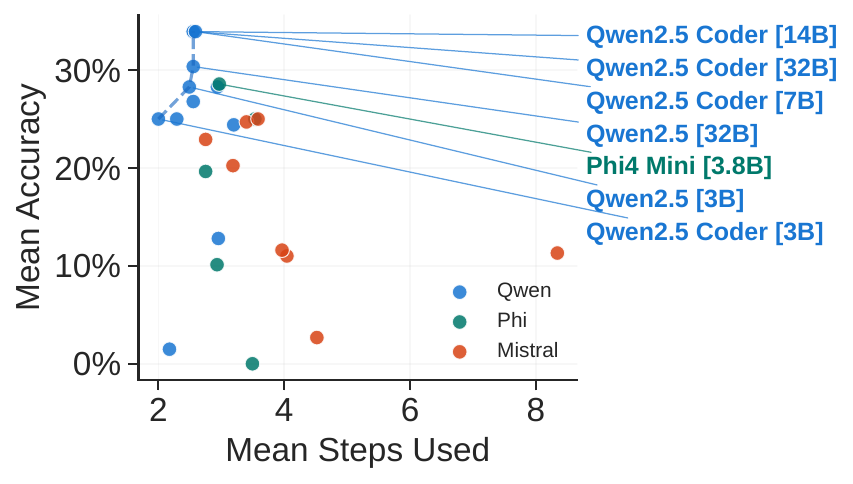}\vspace{-0.5em}
        \caption{Step-accuracy scatter across Qwen, Phi, and Mistral models, with model-name annotations for each point.}
        \label{fig:steps_scatter}
    \end{subfigure}
    \caption{Model-level trade-off views for latency and trajectory length.}
    \label{fig:tradeoff_scatters}
\end{figure*}

\begin{table*}[t]
\caption{Top failure reasons by ecosystem and domain (share of failures), computed over all models in each ecosystem.}
\label{tab:failure_mix}
\centering
\small
\resizebox{0.72\linewidth}{!}{
\begin{tabular}{llrl}
\toprule
\textbf{Model Family} & \textbf{Domain} & \textbf{Mean Accuracy} & \textbf{Failure breakdown} \\
\midrule
\multirow{2}{*}{Qwen} & FinOps & 49.7\% & wrong answer (90.9\%), tool error (4.4\%), max steps reached (2.9\%) \\
\cmidrule{2-4}
  & SRE & 5.9\% & wrong answer (89.3\%), tool error (9.3\%), tool error + max steps reached (1.0\%) \\
\midrule
\multirow{2}{*}{Phi} & FinOps & 35.5\% & tool error + max steps reached (47.9\%), wrong answer (43.0\%), tool error (5.8\%) \\
\cmidrule{2-4}
  & SRE & 1.3\% & tool error + max steps reached (64.9\%), wrong answer (30.1\%), tool error (3.3\%) \\
\midrule
\multirow{2}{*}{Mistral} & FinOps & 33.3\% & wrong answer (58.2\%), tool error (21.5\%), tool error + max steps reached (16.0\%) \\
\cmidrule{2-4}
  & SRE & 2.2\% & wrong answer (52.9\%), tool error + max steps reached (37.4\%), tool error (8.8\%) \\
\bottomrule
\end{tabular}
}
\end{table*}

\subsection{Latency and Step Trade-offs}
To further study deployment trade-offs, Figure~\ref{fig:tradeoff_scatters} presents two complementary model-level views. Figure~\ref{fig:latency_scatter} plots mean accuracy against mean latency. Each point corresponds to one model, grouped by family, and point labels show model names. This view shows that some models deliver only modest accuracy gains at large latency cost, while others occupy better operating points for interactive or resource-constrained settings.
Figure~\ref{fig:steps_scatter} complements latency with the same model-level view against mean step count. This plot helps separate two effects that are often conflated in practice. Some models are slow primarily because each step is expensive, while others consume many tool/model turns to reach a final answer.

Across models, latency and step count are correlated, but not equivalent. We observe settings where step counts are similar while latency differs substantially, indicating that the per-step compute cost dominates. We also observe settings where longer trajectories contribute materially to runtime tails. This means ``cap steps'' and ``choose faster model family'' are complementary controls instead of substitutes.

Taking a closer look, the latency-accuracy scatter in Figure~\ref{fig:latency_scatter} reveals an efficiency pattern among Qwen Coder variants. The model Qwen Coder 7B achieves the same peak accuracy (33.9\%) as its larger 14B and 32B counterparts at substantially lower latency. Specifically, Qwen Coder 7B completes a task in 1.45 seconds on average, compared to 3.35 seconds for Qwen Coder 14B and 6.22 seconds for Qwen Coder 32B. This is a 4.3$\times$ latency reduction for equivalent accuracy. This compression of the efficiency frontier indicates that agentic task quality may not always require proportional latency investment. Careful model selection yields substantial speedups without sacrificing benchmark performance. Beyond the frontier, latency and accuracy are loosely coupled. Phi4 Mini 3.8B reaches only 28.6\% accuracy despite 27-second mean latency, while Qwen 3B achieves 28.3\% at 7.6 seconds. These observations suggest that latency reductions are achievable through strategic model choice, not only through accepting a size-for-accuracy trade-off.

\subsection{Error Patterns}
Failure composition differs sharply by ecosystem and explains several cross-family gaps.

Table~\ref{tab:failure_mix} shows two distinct failure modes. \textit{Semantic failures} (wrong answer) occur when the model completes the trajectory but produces an incorrect final answer, i.e., a reasoning/understanding failure. \textit{Execution failures} (tool error, max steps) occur when the model cannot complete the trajectory, i.e., a tool-use/reasoning-depth failure. 

Qwen is dominated by semantic failures across both domains (89--91\%), indicating that the model can execute the tool workflow but struggles with accurate answer selection. Phi and Mistral show a different pattern, where execution failures dominate, especially for SRE tasks (65\% for Phi, 37\% for Mistral). This suggests that Phi and Mistral models frequently fail to complete the reasoning trajectory instead of producing incorrect answers.

The domain-conditioned view adds another layer. In Phi, SRE tasks have substantially more execution failures (64.9\%) than FinOps (47.9\%), indicating that SRE's multi-source evidence gathering is particularly challenging for this ecosystem. Mistral shows a similar but milder difference. In contrast, Qwen maintains a relatively stable failure composition across domains, with semantic failures dominating in both cases.

These failure profiles have operational implications. 
Qwen's semantic-failure-dominant profile (89--91\% wrong answers) indicates strong tool-execution reliability, i.e., the model consistently completes reasoning trajectories, but final answer selection remains the primary weakness. 
This makes Qwen preferable when trajectory completion and auditability are priorities, with answer verification handled downstream, e.g., by a human. 
In contrast, Phi and Mistral show significant execution failures, particularly on SRE tasks (64.9\% and 37.4\% respectively), meaning these models frequently fail to complete the reasoning trajectory itself. 
While such failures are detectable early and may allow faster fallback triggering, they may represent a more fundamental reliability gap.

\subsection{Domain-Specific Saturation and Bottlenecks}
An important pattern emerges when separating FinOps and SRE within the model-variant experiment. FinOps performance saturates early for many models, while SRE remains difficult even for stronger variants. This divergence suggests the dominant difficulty is less in generic reasoning depth and more in combining multi-source signals and narrowing competing candidates for SRE-style tasks.

This split is operationally significant. If one pipeline mixes cost and reliability diagnostics, a single static model choice is often suboptimal. The same model can perform strongly on FinOps while remaining near floor on SRE.

Together with the error breakdown, this highlights two distinct improvement directions for future systems work: 1) reducing semantic misidentification under rich multi-source context (dominant for SRE), and 2) reducing tool/step-limit failures in smaller instruction-tuned models. Both depend on workflow design and are not solved by parameter scaling alone.

\section{Discussion}
The key message is that edge-agent quality is a systems property, instead of just a parameter-count property. Model family, tool-loop behavior, and task-domain mix jointly determine outcomes.

\noindent\textbf{Non-steady scaling and coder threshold.} Gains from coder-oriented models are highly non-linear, appearing in specific capacity bands instead of increasing smoothly with size. A useful rule of thumb is that coder-oriented models work best after a model already has enough general tool-use reliability. Below that level, they can bring little benefit. This threshold effect explains why some mid-size coder variants match or exceed larger general-purpose models.

\noindent\textbf{Failure-profile-dependent model choice.} Qwen failures are dominated by wrong answers, while Phi and Mistral are dominated by tool/step interaction failures. This implies distinct model-choice preferences under different operational priorities (semantic precision versus execution robustness).

\noindent\textbf{Latency-step trade-offs.} The efficiency patterns reveal substantial savings potential. Qwen Coder 7B achieves the same peak accuracy as its 14B and 32B counterparts at notably lower latency. Some models reach similar accuracy with similar step counts but very different latency. Others are slow because they take longer trajectories. These are different problems, since one asks for faster models, the other for better stopping and tool-path control.

\noindent\textbf{Routing can ease model choice.} The efficiency patterns identified here provide a foundation for automated routing. Recent work shows learned routers achieve substantial cost savings and performance advantages, e.g., RouteLLM achieves 2$\times$ cost reduction without quality loss~\cite{ong2024routellm}, RouterDC uses contrastive learning~\cite{chen2024routerdc}, and MESS+ demonstrates 2$\times$ savings with SLA guarantees~\cite{woisetschlaeger2025mess}. These approaches provide a basis for extending our work to enable agents to pick the right model for the current step.

\noindent\textbf{Domain mix dominates scaling.} The accuracy gap between FinOps and SRE is larger than many within-family gaps. Workload composition can change the preferred model more than parameter count. A team that evaluates only an aggregate score can pick a model that looks strong overall but fails on the tasks that matter most in production. This is especially important for mixed diagnostic pipelines that handle both cost and reliability operations.

\noindent\textbf{Failure composition as primary metric.} For edge settings, stability and predictable failure behavior can matter as much as mean accuracy. Benchmark methodology should treat failure composition (e.g., semantic and execution failures) as a primary evaluation target, not a secondary diagnostic, especially when latency budgets and operational risk constraints are strict.

\noindent\textbf{Practical guidance.} We can conclude the following main lessons based on our empirical study:
\begin{itemize}
    \item Do not assume steady scaling, as mid-size models may match larger ones at lower latency.
    \item Route by workload type, e.g., FinOps and SRE have different difficulty profiles. 
    \item Apply coder-oriented models selectively, as gains appear in specific capacity bands.

\end{itemize}

\section{Conclusion}
We have presented a full-run tool-enabled evaluation of size, generation, and model-variant effects under edge-style constraints. The main findings are as follows: 1) size scaling is non-steady, 2) generation gains come with latency trade-offs, 3) coder versus general-purpose depends on workload, 4) domain gaps are large (FinOps easier than SRE), and 5) failure modes differ. Together, these results support runtime designs that combine model choice, verification, and fallback, instead of relying on parameter scaling alone.

This study is limited to one benchmark setup and a limited set of open models. Future work should apply the same methodology to additional domains (e.g., SecOps), explore hybrid edge-cloud deployment, and investigate how tool-interface design affects model performance.

\balance
\bibliographystyle{ACM-Reference-Format}
\bibliography{references}


\begin{thebibliography}{18}

%%% the Web address.

\ifx \showCODEN    \undefined \def \showCODEN     #1{\unskip}     \fi
\ifx \showISBNx    \undefined \def \showISBNx     #1{\unskip}     \fi
\ifx \showISBNxiii \undefined \def \showISBNxiii  #1{\unskip}     \fi
\ifx \showISSN     \undefined \def \showISSN      #1{\unskip}     \fi
\ifx \showLCCN     \undefined \def \showLCCN      #1{\unskip}     \fi
\ifx \shownote     \undefined \def \shownote      #1{#1}          \fi
\ifx \showarticletitle \undefined \def \showarticletitle #1{#1}   \fi
\ifx \showURL      \undefined \def \showURL       {\relax}        \fi
% invisible to TeX
\providecommand\bibfield[2]{#2}
\providecommand\bibinfo[2]{#2}
\providecommand\natexlab[1]{#1}
\providecommand\showeprint[2][]{arXiv:#2}

\bibitem[Abdin et~al\mbox{.}(2024)]%
        {abdin2024phi3}
\bibfield{author}{\bibinfo{person}{Marah Abdin}, \bibinfo{person}{Jyoti Aneja}, {et~al\mbox{.}}} \bibinfo{year}{2024}\natexlab{}.
\newblock \showarticletitle{Phi-3 Technical Report: A Highly Capable Language Model Locally on Your Phone}.
\newblock \bibinfo{journal}{\emph{arXiv preprint arXiv:2404.14219}} (\bibinfo{year}{2024}).
\newblock


\bibitem[Chen et~al\mbox{.}(2024)]%
        {chen2024routerdc}
\bibfield{author}{\bibinfo{person}{Shuhao Chen}, \bibinfo{person}{Weisen Jiang}, {et~al\mbox{.}}} \bibinfo{year}{2024}\natexlab{}.
\newblock \showarticletitle{RouterDC: Query-Based Router by Dual Contrastive Learning for Assembling Large Language Models}. In \bibinfo{booktitle}{\emph{Advances in Neural Information Processing Systems}}, Vol.~\bibinfo{volume}{37}. \bibinfo{pages}{66305--66328}.
\newblock


\bibitem[Dettmers et~al\mbox{.}(2023)]%
        {dettmers2023qlora}
\bibfield{author}{\bibinfo{person}{Tim Dettmers}, \bibinfo{person}{Artidoro Pagnoni}, {et~al\mbox{.}}} \bibinfo{year}{2023}\natexlab{}.
\newblock \showarticletitle{QLoRA: Efficient Finetuning of Quantized LLMs}. In \bibinfo{booktitle}{\emph{Advances in Neural Information Processing Systems}}, Vol.~\bibinfo{volume}{36}. \bibinfo{pages}{10088--10115}.
\newblock


\bibitem[Jha et~al\mbox{.}(2025)]%
        {jha2025itbench}
\bibfield{author}{\bibinfo{person}{Saurabh Jha}, \bibinfo{person}{Rohan~R. Arora}, {et~al\mbox{.}}} \bibinfo{year}{2025}\natexlab{}.
\newblock \showarticletitle{{ITB}ench: Evaluating {AI} Agents across Diverse Real-World {IT} Automation Tasks}. In \bibinfo{booktitle}{\emph{Proceedings of the 42nd International Conference on Machine Learning}} \emph{(\bibinfo{series}{Proceedings of Machine Learning Research}, Vol.~\bibinfo{volume}{267})}. \bibinfo{pages}{27134--27197}.
\newblock


\bibitem[{Kaggle}(2026)]%
        {kaggle2026itbench}
\bibfield{author}{\bibinfo{person}{{Kaggle}}.} \bibinfo{year}{2026}\natexlab{}.
\newblock \bibinfo{title}{ITBench Benchmark Leaderboard}.
\newblock \bibinfo{howpublished}{\url{https://www.kaggle.com/benchmarks/ibm-research/itbench}}.
\newblock


\bibitem[Kojima et~al\mbox{.}(2022)]%
        {kojima2022zeroshot}
\bibfield{author}{\bibinfo{person}{Takeshi Kojima}, \bibinfo{person}{Shixiang~Shane Gu}, {et~al\mbox{.}}} \bibinfo{year}{2022}\natexlab{}.
\newblock \showarticletitle{Large Language Models are Zero-Shot Reasoners}. In \bibinfo{booktitle}{\emph{Advances in Neural Information Processing Systems}}, Vol.~\bibinfo{volume}{35}. \bibinfo{pages}{22199--22213}.
\newblock


\bibitem[Liu et~al\mbox{.}(2024)]%
        {liu2024agentbench}
\bibfield{author}{\bibinfo{person}{Xiao Liu}, \bibinfo{person}{Hao Yu}, {et~al\mbox{.}}} \bibinfo{year}{2024}\natexlab{}.
\newblock \showarticletitle{AgentBench: Evaluating {LLM}s as Agents}. In \bibinfo{booktitle}{\emph{The Twelfth International Conference on Learning Representations}}.
\newblock
\urldef\tempurl%
\url{https://openreview.net/forum?id=zAdUB0aCTQ}
\showURL{%
\tempurl}


\bibitem[Madaan et~al\mbox{.}(2023)]%
        {madaan2023selfrefine}
\bibfield{author}{\bibinfo{person}{Aman Madaan}, \bibinfo{person}{Niket Tandon}, {et~al\mbox{.}}} \bibinfo{year}{2023}\natexlab{}.
\newblock \showarticletitle{Self-Refine: Iterative Refinement with Self-Feedback}. In \bibinfo{booktitle}{\emph{Advances in Neural Information Processing Systems}}, Vol.~\bibinfo{volume}{36}. \bibinfo{pages}{46534--46594}.
\newblock


\bibitem[Mialon et~al\mbox{.}(2023)]%
        {mialon2023gaia}
\bibfield{author}{\bibinfo{person}{Gr{\'e}goire Mialon}, \bibinfo{person}{Cl{\'e}mentine Fourrier}, {et~al\mbox{.}}} \bibinfo{year}{2023}\natexlab{}.
\newblock \showarticletitle{GAIA: a benchmark for General AI Assistants}.
\newblock \bibinfo{journal}{\emph{arXiv preprint arXiv:2311.12983}} (\bibinfo{year}{2023}).
\newblock


\bibitem[{NVIDIA}(2024)]%
        {nvidia2024jetsonnano}
\bibfield{author}{\bibinfo{person}{{NVIDIA}}.} \bibinfo{year}{2024}\natexlab{}.
\newblock \bibinfo{title}{Jetson Nano Developer Platform}.
\newblock \bibinfo{howpublished}{\url{https://developer.nvidia.com/embedded/jetson-nano}}.
\newblock


\bibitem[Ong et~al\mbox{.}(2025)]%
        {ong2024routellm}
\bibfield{author}{\bibinfo{person}{Isaac Ong}, \bibinfo{person}{Amjad Almahairi}, {et~al\mbox{.}}} \bibinfo{year}{2025}\natexlab{}.
\newblock \showarticletitle{Route{LLM}: Learning to Route {LLM}s from Preference Data}. In \bibinfo{booktitle}{\emph{The Thirteenth International Conference on Learning Representations}}.
\newblock
\urldef\tempurl%
\url{https://openreview.net/forum?id=8sSqNntaMr}
\showURL{%
\tempurl}


\bibitem[{OpenAI}(2023)]%
        {openai2023gpt4}
\bibfield{author}{\bibinfo{person}{{OpenAI}}.} \bibinfo{year}{2023}\natexlab{}.
\newblock \showarticletitle{{GPT}-4 Technical Report}.
\newblock \bibinfo{journal}{\emph{arXiv preprint arXiv:2303.08774}} (\bibinfo{year}{2023}).
\newblock


\bibitem[Wang et~al\mbox{.}(2024)]%
        {wang2024officebench}
\bibfield{author}{\bibinfo{person}{Zilong Wang}, \bibinfo{person}{Yuedong Cui}, {et~al\mbox{.}}} \bibinfo{year}{2024}\natexlab{}.
\newblock \showarticletitle{OfficeBench: Benchmarking Language Agents across Multiple Applications for Office Automation}.
\newblock \bibinfo{journal}{\emph{arXiv preprint arXiv:2407.19056}} (\bibinfo{year}{2024}).
\newblock


\bibitem[Wei et~al\mbox{.}(2022)]%
        {wei2022cot}
\bibfield{author}{\bibinfo{person}{Jason Wei}, \bibinfo{person}{Xuezhi Wang}, {et~al\mbox{.}}} \bibinfo{year}{2022}\natexlab{}.
\newblock \showarticletitle{Chain-of-Thought Prompting Elicits Reasoning in Large Language Models}. In \bibinfo{booktitle}{\emph{Advances in Neural Information Processing Systems}}, Vol.~\bibinfo{volume}{35}. \bibinfo{pages}{24824--24837}.
\newblock


\bibitem[Woisetschl{\"a}ger et~al\mbox{.}(2025)]%
        {woisetschlaeger2025mess}
\bibfield{author}{\bibinfo{person}{Herbert Woisetschl{\"a}ger}, \bibinfo{person}{Ryan Zhang}, {et~al\mbox{.}}} \bibinfo{year}{2025}\natexlab{}.
\newblock \showarticletitle{{MESS}+: Dynamically Learned Inference-Time {LLM} Routing in Model Zoos with Service Level Guarantees}. In \bibinfo{booktitle}{\emph{The Thirty-ninth Annual Conference on Neural Information Processing Systems}}.
\newblock
\urldef\tempurl%
\url{https://openreview.net/forum?id=wIM0y07NGX}
\showURL{%
\tempurl}


\bibitem[Yao et~al\mbox{.}(2023)]%
        {yao2023react}
\bibfield{author}{\bibinfo{person}{Shunyu Yao}, \bibinfo{person}{Jeffrey Zhao}, {et~al\mbox{.}}} \bibinfo{year}{2023}\natexlab{}.
\newblock \showarticletitle{ReAct: Synergizing Reasoning and Acting in Language Models}. In \bibinfo{booktitle}{\emph{The Eleventh International Conference on Learning Representations}}.
\newblock
\urldef\tempurl%
\url{https://openreview.net/forum?id=WE_vluYUL-X}
\showURL{%
\tempurl}


\bibitem[Yuan et~al\mbox{.}(2025)]%
        {yuan2025localcloud}
\bibfield{author}{\bibinfo{person}{Liangqi Yuan}, \bibinfo{person}{Dong-Jun Han}, {et~al\mbox{.}}} \bibinfo{year}{2025}\natexlab{}.
\newblock \showarticletitle{Local-Cloud Inference Offloading for LLMs in Multi-Modal, Multi-Task, Multi-Dialogue Settings}. In \bibinfo{booktitle}{\emph{Proceedings of the Twenty-Sixth International Symposium on Theory, Algorithmic Foundations, and Protocol Design for Mobile Networks and Mobile Computing}} \emph{(\bibinfo{series}{MobiHoc '25})}. \bibinfo{pages}{201–210}.
\newblock


\bibitem[Zhang et~al\mbox{.}(2024)]%
        {zhang2024tinyllama}
\bibfield{author}{\bibinfo{person}{Peiyuan Zhang}, \bibinfo{person}{Guangtao Zeng}, {et~al\mbox{.}}} \bibinfo{year}{2024}\natexlab{}.
\newblock \showarticletitle{TinyLlama: An Open-Source Small Language Model}.
\newblock \bibinfo{journal}{\emph{arXiv preprint arXiv:2401.02385}} (\bibinfo{year}{2024}).
\newblock


\end{thebibliography}

\end{document}